\documentclass[sigconf]{acmart}
\usepackage{microtype}
\usepackage{graphicx}
\usepackage{subfigure}
\usepackage{subcaption}
\usepackage{array, makecell}
\usepackage{booktabs} 
\usepackage{hyperref}
\usepackage{multirow}
\usepackage{geometry}
\AtBeginDocument{%
  }

\begin{document}

\title{Investigating Context Effects in Similarity Judgements in Large Language Models}

\author{Sagar Uprety}
\email{s.uprety@ucl.ac.uk}
\affiliation{%
  \institution{University College London, United Kingdom }
  \country{United Kingdom}
  }
\author{Amit Kumar Jaiswal}
\email{a.jaiswal@surrey.ac.uk}
\affiliation{%
  \institution{University of Surrey, United Kingdom}
  \country{United Kingdom}
}

\author{Haiming Liu}
\email{h.liu@soton.ac.uk}
\affiliation{%
  \institution{University of Southampton, United Kingdom}
  \country{United Kingdom}
  }

\author{Dawei Song}
\email{dawei.song@open.ac.uk}
\affiliation{%
  \institution{The Open University, United Kingdom}
  \country{United Kingdom}
}

\renewcommand{\shortauthors}{Uprety et al.}

\begin{abstract}
Large Language Models (LLMs) have revolutionised the capability of AI models in comprehending and generating natural language text. They are increasingly being used to empower and deploy agents in real-world scenarios, which make decisions and take actions based on their understanding of the context. Therefore researchers, policy makers and enterprises alike are working towards ensuring that the decisions made by these agents align with human values and user expectations. That being said, human values and decisions are not always straightforward to measure and are subject to different cognitive biases. There is a vast section of literature in Behavioural Science which studies biases in human judgements. In this work we report an ongoing investigation on alignment of LLMs with human judgements affected by order bias.  Specifically, we focus on a famous human study which showed evidence of order effects in similarity judgements, and replicate it with various popular LLMs. We report the different settings where LLMs exhibit human-like order effect bias and discuss the implications of these findings to inform the design and development of LLM based applications.
\end{abstract}


\maketitle

\section{Introduction}
\label{Introduction}
Large Language models (LLMs) like GPT-4~\cite{gpt-4} and LLaMa family~\cite{llama2} have been at the forefront of ushering a new wave of research, funding, investments and public opinion in Artificial Intelligence (AI). This is due by their remarkably improved capabilities in reasoning~\cite{wei2022chain_of_thought, huang2023reasoning_survey}, cross-domain generalisation~\cite{10.5555/llm_few_shot_learners} and the emergence of a new type of learning - in-context learning~\cite{wei2022emergent_abilities_llms}. Some researchers have gone as far as suggesting that LLMs are the pre-cursor to a generalist form of AI~\cite{bubeck2023sparks_of_agi}, whereas there is another school of thought which says that the autoregressive training and next word prediction architecture of LLMs are not enough to imitate human cognitive intelligence~\cite{schaeffer2024are_emergent_ab_mirage}. Nevertheless, it is accepted that LLMs are useful tools to augment our workflows and assist in different tasks like coding assistant\footnote{https://github.com/features/copilot}, copywriting, marketing and sales\footnote{https://www.jasper.ai, https://www.copy.ai}, systematic reviews\footnote{https://elicit.com/, https://scholarai.io/}, search engines\footnote{https://you.com/, https://www.perplexity.ai/}, etc. There is ongoing research in developing specialised and generalist LLM agents which can use other tools like browsers and APIs\footnote{https://www.multion.ai/} to perform tasks like booking holidays, reserving tables, online shopping, etc. Increasingly, many of our daily workflows, including our professional work will be either automated or augmented with the help of LLM-based agents. These agents will interact with and impact the society in general.

Thus, it is imperative that the decisions made by these agents through LLMs align with not only the intentions of the users of these agent applications, but also societal values and conventions. There is a parallel line of research in this direction under the umbrella of AI Alignment~\cite{ji2023ai_alignment, gabriel2020ai_alignment}.

There are several facets of AI alignment, depending upon the application domain of the AI model. However, the underlying theme of current alignment research is the focus on avoiding negative or incorrect outcomes from LLMs and also the focus on \textbf{values}~\cite{rane2024concept_align}. There is a large class of human judgements where the outcome is neither negative nor incorrect, rather there is no single correct outcome. This is because the outcome depends on the context of judgement. 
In this work, we focus on the alignment of context-sensitive judgements produced by LLMs with those produced by humans. There are several implications of aligning human-LLM judgements and decisions. Firstly, it can help test theories in psychology and cognitive science on large scale human-like data~\cite{bhatia2023inductive, bhatia2022transformer, bhatia2024exploring}. Secondly, it will help improve performance of LLM agents in real-world scenarios by making them robust to change in contexts and uncertainty. We focus on a certain kind of judgements - similarity judgement, which underpins a wide range of applications such as search,  question answering and recommendation. The ability to judge similarity of stimuli, concepts, entities, etc. is an important aspect decision-making in humans~\cite{Quine1969-QUIORA-6}. Our research question is :

\textbf{(RQ): How well do LLMs align with humans in context-sensitive similarity judgements?
}

We replicate a study by Tversky and Gati~\cite{Tversky_gati_similarity}, which was one of several which investigated violation of symmetry in human similarity judgements. Similarity of concepts and entities is mathematically represented using distance metrics in some coordinate space. One of the chief properties of such spaces is that metric distance or vector projections are symmetric, i.e. the similarity of two entities A and B is same as similarity between B and A. A large array of work from Amol Tversky~\cite{Tversky1977FeaturesOS, Tversky_gati_similarity} has investigated and found that similarity judgements lose their symmetric property when made under context-sensitive scenarios. The concepts and entities considered in these studies have ranged from country-pairs to geometric figures. In the present work, we use the study conducted to judge similarity of pairs of countries where the context is created by the order of the country in the pair. For e.g. one group of participants asked to rate the similarity of China and North Korea and another group is asked to rate the similarity of North Korea and China. Our hypothesis is that LLMs will also exhibit human-like assymmetric judgements because of the change of context exhibited by changing the order of countries in a pair. We also conduct a novel variant of the experiment by prompting the LLMs with both orders of countries in the same prompt. This is akin to asking a human participant to rate the similarity of countries in both orders. Most humans should give the exact same score in such a case, as the concept of similarity is intuitively understood to be symmetric. Similarly, we hypothesise that LLMs should also render the exact same score for the two different orders of the country pairs. 

We study eight different LLMs - both closed-source and open-source, small and large in their number of parameters. We utilise different temperature settings and different variants of prompts in terms of the text. Like human judgements, LLMs are also found to produce different ratings for similarity based on the order of countries in the pair, however not all such instances are statistically significant. We report on the model-temperature pairs which show statistically significant order effects, discuss the implications of these results and discuss the future steps in this line of research.



\vspace{-2mm}

\section{Related work} 
Research at the intersection of cognitive science and LLMs is based on chiefly two perspectives. Although both perspectives seek to know how close are LLMs to humans, one line of work is interested in utilising LLMs to study cognitive models and theories on a large scale by replicating human behaviour with synthetic data generated from LLMs~\cite{bhatia2022transformer, bhatia2023inductive, bhatia2024exploring}. The other research direction concerns with the application of LLMs in automation and augmentation, wherein it is essential that decisions and judgements by LLM based agents align with human preferences and judgements~\cite{almeida2024exploring_psych, irrationality_survey_macmillan2023ir}. In either case, while comparing humans and LLMs in terms of responses, judgements and decisions, it is important to know whether the underlying mechanisms of representation of concepts and reasoning are similar or not. In general, it is seen that even in cases where LLM judgements are similar to humans, the underlying processes of reasoning and representation are different \cite{bhatia2023inductive, irrationality_macmillan2024ir}. For e.g.,
~\cite{bhatia2024exploring} reveals that while LLMs are effective at understanding and explaining variability in human decisions under risk, their risk assessments are purely data-driven, lacking the emotional and psychological factors influencing human decisions. 


A large array of recent work investigates presence of different human-like cognitive biases in LLMs~\cite{echterhoff2024cognitive_biases_LLMs, cheunglarge_amplify_human_biases, irrationality_macmillan2024ir, alsagheer2024comparing_irrationality}. In ~\cite{zhu2024incoherent}, the authors repeatedly prompt LLMs with the same query and study the variability in the responses. It is found that at temperature = $1.0$, the variability displays similar patterns to human judgements variability. In ~\cite{irrationality_macmillan2024ir} and \cite{alsagheer2024comparing_irrationality}, LLMs are tested against popular tasks like Wason selection and Conjunction fallacy and found to contain human-like bias in some of the tasks. Another important study ~\cite{echterhoff2024cognitive_biases_LLMs} found presence of common cognitive biases like anchoring, framing, group atttribution in 4 LLMs - GPT-3.5-Turbo, GPT-4, Llama2 7B, and Llama2 13B. However, they were able to devise strategies to mitigate these biases to a good extent. 

Human judgements typically optimise over a lot of different factors thus leading to a lot of trade-offs (Bounded rationality). One such trade-off between truthfulness and helpfulness to user goals of LLM responses is studied in ~\cite{liu2024large}. This balance is crucial because, like humans, LLMs must sometimes prioritise helpfulness over strict honesty to be effective communicators. For instance, an LLM might simplify or approximate numerical values to aid understanding, akin to how humans might round numbers in conversation to make information more comprehensible. It was found that LLM responses also display human-like patterns, which can be further tuned towards either factor using different techniques like RLHF or Chain-of-Though Prompting (CoT).

\vspace{-2mm}

\section{Tversky and Gati's Similarity Effects Study}

In this study~\cite{Tversky_gati_similarity}, participants were shown pairs of $21$ countries and asked to rate the similarity between them on a scale of $0$ to $20$. The participants were divided into two groups, and each group rated similarity of $21$ countries. The only difference between the groups was the order of countries in the pair. It was found that there were statistically significant differences in similarity scores for the country pairs in the two groups, thus providing evidence that human given similarity judgements may not always follow the symmetry assumption. 

Tversky and Gati consider similarity judgements to be an extension of similarity statements (\textit{a is like b}), which are in general - directional. When two entities are compared, one is usually the subject and the other is the referent. The referent is always the more prominent of the two entities (i.e. has with more salient features). In the example given earlier in the paper, China would be the referent and North Korea the subject. Tversky and Gati hypothesise that if $s(a,b)$ and $s(b,a)$ are the similarity of a pair of subject and referent in the two orders, then $s(a,b) > s(b,a)$. Similarity of two entities would be more if the first entity is the less prominent one of the pair. This is validated in their experiment for the $21$ country pairs when they find that a one-sided paired t-test leads to statistically significant improvement in similarity scores when a less prominent country (in terms of size, or economic and political influence) is placed in the first position in a pair.

One possible explanation of difference in similarity scores because of different order of presentation is that different orders lead to different factors used  by participants to evaluate the similarity of the two countries. E.g., in case of North Korea and China, when presented in this order, participants consider specific factors like political system, culture, proximity as some of the factors which are common to both countries. However, when people come across 'China' as the first country in the pair, the factors of size, economy, etc. come into consideration. Across these factors North Korea doesn't seem much similar to China and thus people give a lower score. Thus, a more prominent country, when considered first, has many more features, thereby leading to a small fraction overlapping with those of a smaller country. While this explanation sounds intuitive, it violates the symmetric property of distance-based similarity measures.
\vspace{-0mm}

\section{Experiments}
\begin{table}[tb]
\small
\caption{T-test results for various prompt styles, models, and temperatures}\label{tab:t_test_results}
\vspace{-10pt}
\begin{tabular}{|c|c|c|c|c|c|c|}
\hline
\textbf{Prompt} & \textbf{Model} & \textbf{Temp.}  & \textbf{t-stat} & \textbf{p-val} & \textbf{Dual setting} \\
\hline
SSA & mistral\_7B & 1.0  & -2.107 & \textbf{0.024} & p-val = 0.859\\
SSA & mistral\_7B & 1.5  & -1.775 & \textbf{0.046} & p-val= 0.391\\
SST & \textbf{llama3\_8B} & \textbf{0.001}  & \textbf{-2.149} & \textbf{0.022} & \textbf{Symmetric}\\
SST & \textbf{llama3\_8B} & \textbf{0.5}  & \textbf{-2.251} & \textbf{0.018}  & \textbf{Symmetric}\\
SSD & llama2\_13B & 0.5 &  -1.896 & \textbf{0.036} & p-val = 0.773\\
SSD & llama2\_13B & 1.5  & -1.936 & \textbf{0.034} & p-val = 0.848\\
SSA & llama2\_70B & 0.5  & -2.278 & \textbf{0.017} & p-val = 0.051\\
SSD & \textbf{gpt-4 }& \textbf{0.5} & \textbf{-1.917} &\textbf{ 0.035} & \textbf{Symmetric} \\ \hline
\end{tabular}
\vspace{-5mm}
\end{table}

\subsection{Models}
We conduct investigations into order effect in similarity using 8 LLMs - Mistral 7B\footnote{\url{https://docs.mistral.ai/getting-started/models/}}, Llama2 7B, Llama2 13B, LLama2 70B, Llama3 8B, Llama3 70B~\cite{llama2}, GPT-3.5 and GPT-4~\cite{gpt-4}. For the open-source models we use their non-quantised 16 bit versions, with default values of top\_p ($0.9$) and top\_k ($50$) parameters, while the temperature parameter is varied. The different temperature parameters used are $0.001, 0.5, 1.0, 1.5$. Note that we did not use temperature = $0$, as the Llama models require a non-zero temperature value, unlike OpenAI models. We thus keep the lowest temperature value to as low as possible, and avoid using temperature = $0$ just for the OpenAI models. 

The prompts for each of the LLMs consist of a system message - which also instructs the model to generate response in a structured json format, and the specific questions which are part of the study. 
\vspace{-2mm}

\subsection{Experiment 1 - Single pair prompts}
In the original study the participants were asked to "assess the degree to which country A similar to country B". Therefore, the equivalent prompt to this question is "On a scale of 0 to 20, where 0 means no similarity and 20 means complete similarity, assess the degree to which \{country\_1\} is similar to \{country\_2\}?". Moreover, each prompt is preceded by a system message and we assert the LLM to strictly return a number between $0$ and $20$ in a json format. However, apart from this prompt, we also try two other variants by changing the language of the prompt. It is well-known that LLM outputs, even at zero temperature, are sensitive to choice of words in a prompt~\cite{lu2021fantastically_prompt_order}. The expected output is a json of shape: \{'score':  \}. The three variants of single pair prompts with appropriate labels are:

\begin{itemize}
    \item \textbf{Single Similar Degree (SSD)} - On a scale of 0 to 20, where 0 means no similarity and 20 means complete similarity, assess the degree to which \{country\_1\} is similar to \{country\_2\}?

    \item \textbf{Single Similar To (SST)} - On a scale of 0 to 20, where 0 means no similarity and 20 means complete similarity, is \{country\_1\}  similar to \{country\_2\} ? 

    \item \textbf{Single Similar And (SSA)} - On a scale of 0 to 20, where 0 means no similarity and 20 means complete similarity, how similar are \{country\_1\} and \{country\_2\}?
\end{itemize}
In the other order of these prompts, the country names are swapped.
\vspace{-2mm}
\vspace{-2mm}

\subsection{Experiment 2 - Dual pair prompts}
In this experimental setting we prompt the LLMs with both the orders of the two countries present in the same prompt, thus eliminating the context created by changing orders. Although there will always be a particular order in which the countries are presented, if the other order immediately follows the first order, the context effect is negated. 

Each prompt consist of two statements for each order. The expected output is a json with two keys - score\_1 and score\_2. We again use the three different wording styles in both statements. The three prompts with their labels are:

\begin{itemize}
    \item \textbf{Dual Similar Degree (DSD)} -  
    
    \textbf{Question 1:} On a scale of 0 to 20, where 0 means no similarity and 20 means complete similarity, assess the degree to which \{country\_1\} is similar to \{country\_2\}?
    
    \textbf{Question 2:} On a scale of 0 to 20, where 0 means no similarity and 20 means complete similarity, assess the degree to which \{country\_2\} is similar to \{country\_1\}? 

    \item \textbf{Dual Similar To (DST) }- 
    
    \textbf{Question 1:} On a scale of 0 to 20, where 0 means no similarity and 20 means complete similarity, is \{country\_1\}  similar to \{country\_2\}?
    
  \textbf{Question 2:} On a scale of 0 to 20, where 0 means no similarity and 20 means complete similarity, is \{country\_2\}  similar to \{country\_1\}?

    \item \textbf{Dual Similar And (DSA)} -
    
    \textbf{Question 1:} On a scale of 0 to 20, where 0 means no similarity and 20 means complete similarity, how similar are \{country\_1\} and \{country\_2\}? 
    
     \textbf{Question 2:} On a scale of 0 to 20, where 0 means no similarity and 20 means complete similarity, how similar are \{country\_2\} and \{country\_1\}? 
\end{itemize}

\begin{figure*}[h!]
  \centering
    \subfigure[]{\includegraphics[scale=0.27]{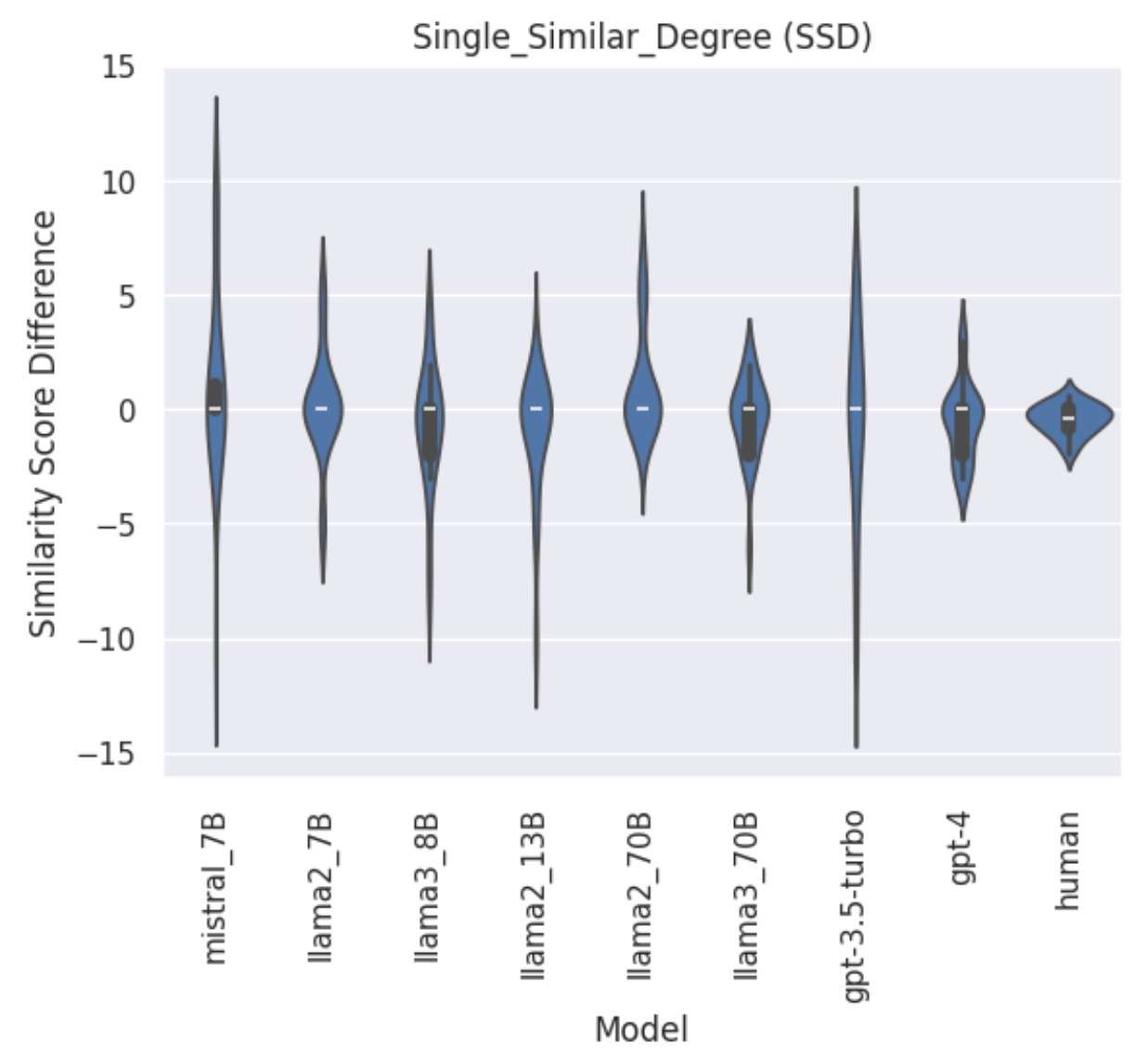}}
  \subfigure[]{\includegraphics[scale=0.27]{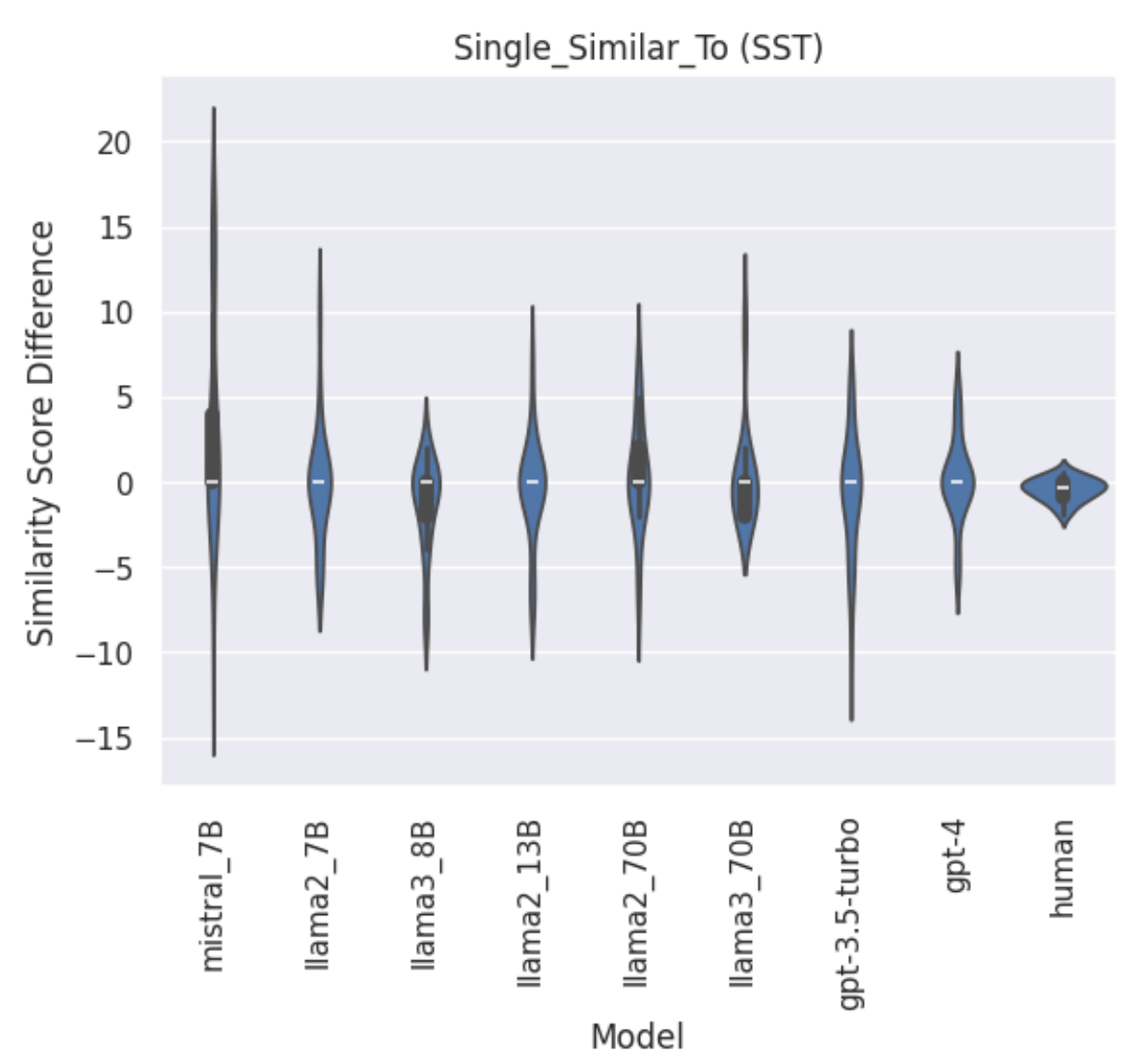}}
    \subfigure[]{\includegraphics[scale=0.27]{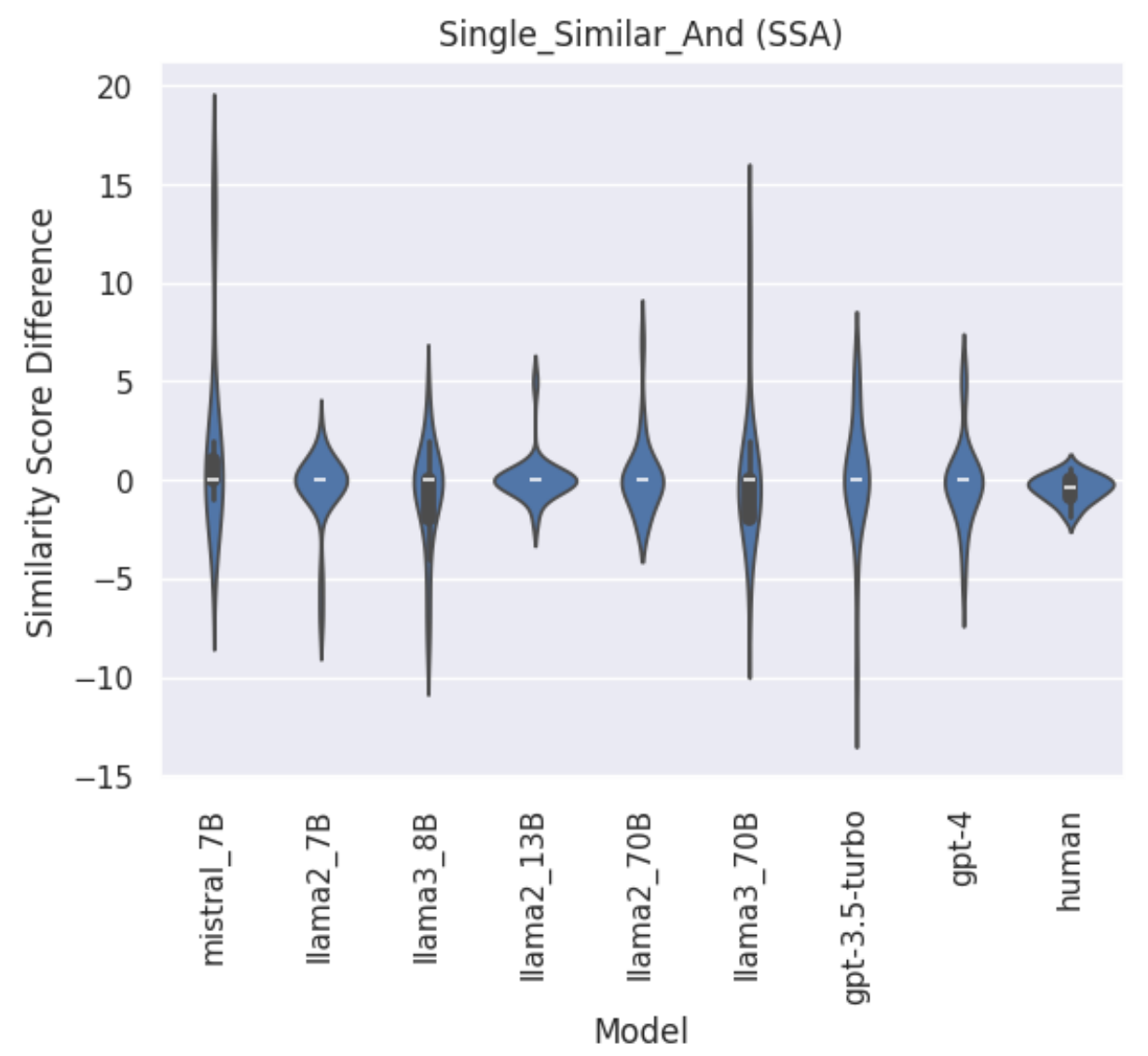}}
    \caption{Distribution of similarity differences for all countries and all models for single prompt settings}\label{fig:distribution-plot-single}
\end{figure*}

\vspace{-0mm}

\section{Results and Discussion}

We perform the above experiments for all $21$ pairs of countries, for all $8$ models, $4$ temperature settings per model and $3$ different single-prompt styles, totalling in $96$ different instances of models-temperature-prompt settings for the set of $21$ country pairs. Figure \ref{fig:distribution-plot-single} shows the distribution of difference in scores for all country pairs for all models, temperature settings and single prompt styles. We also plot the human similarity score differences from Tversky and Gati's study. It is evident that all models show asymmetry in similarity judgements, a departure from the mathematical definition. However, as will be detailed below, not all settings have statistically significant differences. For the dual prompt setting, the hypothesis is that all similarity judgements should be symmetrical, as when deliberately shown both orders, humans should give the same rating to the country pairs. The similarity differences for dual prompt settings are plotted in figure \ref{fig:distribution-plot-dual}.

Similar to ~\cite{Tversky_gati_similarity}, an order effect is calculated when the mean difference between the similarity scores is statistically significant ($\alpha < 0.05$) based on a one-sided paired t-test. The direction being the increase in similarity score when the less prominent country is placed at first in the pair. According to our second alignment criteria, the scores across the dual prompt styles should be symmetric for all models. Of the $96$ different settings, we find only 9 settings where \textbf{both} the criteria are met. Table \ref{tab:t_test_results} lists all these settings. We list the t-statistic and p-value for each model-temperature-single prompt style setting. We also check the effects in the corresponding dual prompt version for the same model-temperature setting. Either there are no similarity score differences for all $21$ countries in the two orders for that setting, or the differences are not statistically significant. These could be due to other reasons such as prompt sensitivity~\cite{lu2021fantastically_prompt_order}. 

 We can see that only 3 models have perfectly symmetric similarity scores for all the dual prompt styles and temperature settings - Llama3 8B, Llama3 70B and GPT-4. There are three instances of models scoring perfectly symmetrically in the dual prompt settings - GPT-4 for the original study's SSD prompt style and Llama3\_8B for the SST prompt style for two different temperature settings.  We can thus say that GPT-4 is aligned with human judgements for this similarity task (for temperature = $0.5$) as the SSD prompt is closest to the original study setting. However, the order effect in the original study is not hypothesised to be due to the wording of the question itself, but rather due to the order of the countries in the pair. It is safe to assume that if the questions in the original study were of style SST, the order effects would have still been present in the similarity judgements. Therefore, we can say that Llama3\_8B model also aligns with human judgements for this task. Moreover, it is aligned for two temperature values versus one value for GPT-4. 

However, largely the LLMs do not show significant asymmetry in similarity judgements. Only some models with certain temperature settings do so. This means that LLMs can be made to align their behaviour with human behaviour by tuning these settings. There might be certain applications where this alignment is beneficial and certain where it is not. For example, consider an e-commerce LLM-based chat agent asked by whether Phone A is similar to Phone B or not. If another user asks the agent the same question in another order (perhaps they are more inclined towards the other phone), the agent should ideally come up with the same response. Should the LLM exhibit order effects in similarity judgements, it would report inconsistent results to different users solely based on the order of items in query. On the other hand, there might be certain applications e.g. in dating, mental health support where LLMs might be better-off aligned to human order bias.

\vspace{-4mm}

\section{Conclusion \& Future Work}
In this paper, we explored the how aligned are LLM judgements with human judgements via similarity ratings to pairs of countries in different orders. Our investigation involves studying the effect of context-driven asymmetry in human judgements utilising Tversky's framework of similarity judgements~\cite{Tversky_gati_similarity}. We find that out of eight LLMs studied across different temperature settings and prompting styles, only Llama3 8B and GPT-4 models report statistically significant order effects in line with human judgements. Even with these models, changing the temperature setting can lead to disappearance of the effect. In the future, we aim to elicit and compare the reasoning generated by these models by different prompting approaches like Chain of Thought prompting~\cite{wei2022chain_of_thought}. We can then compare the crtieria used by LLMs to arrive at their similarity score and whether and how this criteria is different for different order of countries in the pair.

\bibliographystyle{ACM-Reference-Format}
\bibliography{bibliography}


\begin{thebibliography}{26}


\ifx \showCODEN    \undefined \def \showCODEN     #1{\unskip}     \fi
\ifx \showDOI      \undefined \def \showDOI       #1{#1}\fi
\ifx \showISBNx    \undefined \def \showISBNx     #1{\unskip}     \fi
\ifx \showISBNxiii \undefined \def \showISBNxiii  #1{\unskip}     \fi
\ifx \showISSN     \undefined \def \showISSN      #1{\unskip}     \fi
\ifx \showLCCN     \undefined \def \showLCCN      #1{\unskip}     \fi
\ifx \shownote     \undefined \def \shownote      #1{#1}          \fi
\ifx \showarticletitle \undefined \def \showarticletitle #1{#1}   \fi
\ifx \showURL      \undefined \def \showURL       {\relax}        \fi
\providecommand\bibfield[2]{#2}
\providecommand\bibinfo[2]{#2}
\providecommand\natexlab[1]{#1}
\providecommand\showeprint[2][]{arXiv:#2}

\bibitem[Achiam et~al\mbox{.}(2023)]%
        {gpt-4}
\bibfield{author}{\bibinfo{person}{Josh Achiam}, \bibinfo{person}{Steven Adler}, \bibinfo{person}{Sandhini Agarwal}, \bibinfo{person}{Lama Ahmad}, \bibinfo{person}{Ilge Akkaya}, \bibinfo{person}{Florencia~Leoni Aleman}, \bibinfo{person}{Diogo Almeida}, \bibinfo{person}{Janko Altenschmidt}, \bibinfo{person}{Sam Altman}, \bibinfo{person}{Shyamal Anadkat}, {et~al\mbox{.}}} \bibinfo{year}{2023}\natexlab{}.
\newblock \showarticletitle{Gpt-4 technical report}.
\newblock \bibinfo{journal}{\emph{arXiv preprint arXiv:2303.08774}} (\bibinfo{year}{2023}).
\newblock


\bibitem[Almeida et~al\mbox{.}(2024)]%
        {almeida2024exploring_psych}
\bibfield{author}{\bibinfo{person}{Guilherme~FCF Almeida}, \bibinfo{person}{Jos{\'e}~Luiz Nunes}, \bibinfo{person}{Neele Engelmann}, \bibinfo{person}{Alex Wiegmann}, {and} \bibinfo{person}{Marcelo de Ara{\'u}jo}.} \bibinfo{year}{2024}\natexlab{}.
\newblock \showarticletitle{Exploring the psychology of LLMs’ Moral and Legal Reasoning}.
\newblock \bibinfo{journal}{\emph{Artificial Intelligence}} (\bibinfo{year}{2024}), \bibinfo{pages}{104145}.
\newblock


\bibitem[Alsagheer et~al\mbox{.}(2024)]%
        {alsagheer2024comparing_irrationality}
\bibfield{author}{\bibinfo{person}{Dana Alsagheer}, \bibinfo{person}{Rabimba Karanjai}, \bibinfo{person}{Nour Diallo}, \bibinfo{person}{Weidong Shi}, \bibinfo{person}{Yang Lu}, \bibinfo{person}{Suha Beydoun}, {and} \bibinfo{person}{Qiaoning Zhang}.} \bibinfo{year}{2024}\natexlab{}.
\newblock \showarticletitle{Comparing Rationality Between Large Language Models and Humans: Insights and Open Questions}.
\newblock \bibinfo{journal}{\emph{arXiv preprint arXiv:2403.09798}} (\bibinfo{year}{2024}).
\newblock


\bibitem[Bhatia(2023)]%
        {bhatia2023inductive}
\bibfield{author}{\bibinfo{person}{Sudeep Bhatia}.} \bibinfo{year}{2023}\natexlab{}.
\newblock \showarticletitle{Inductive reasoning in minds and machines.}
\newblock \bibinfo{journal}{\emph{Psychological Review}} (\bibinfo{year}{2023}).
\newblock


\bibitem[Bhatia(2024)]%
        {bhatia2024exploring}
\bibfield{author}{\bibinfo{person}{Sudeep Bhatia}.} \bibinfo{year}{2024}\natexlab{}.
\newblock \showarticletitle{Exploring variability in risk taking with large language models.}
\newblock \bibinfo{journal}{\emph{Journal of Experimental Psychology: General}} (\bibinfo{year}{2024}).
\newblock


\bibitem[Bhatia and Richie(2022)]%
        {bhatia2022transformer}
\bibfield{author}{\bibinfo{person}{Sudeep Bhatia} {and} \bibinfo{person}{Russell Richie}.} \bibinfo{year}{2022}\natexlab{}.
\newblock \showarticletitle{Transformer networks of human conceptual knowledge.}
\newblock \bibinfo{journal}{\emph{Psychological Review}} (\bibinfo{year}{2022}).
\newblock


\bibitem[Brown et~al\mbox{.}(2020)]%
        {10.5555/llm_few_shot_learners}
\bibfield{author}{\bibinfo{person}{Tom~B. Brown}, \bibinfo{person}{Benjamin Mann}, \bibinfo{person}{Nick Ryder}, \bibinfo{person}{Melanie Subbiah}, \bibinfo{person}{Jared Kaplan}, \bibinfo{person}{Prafulla Dhariwal}, \bibinfo{person}{Arvind Neelakantan}, \bibinfo{person}{Pranav Shyam}, \bibinfo{person}{Girish Sastry}, \bibinfo{person}{Amanda Askell}, \bibinfo{person}{Sandhini Agarwal}, \bibinfo{person}{Ariel Herbert-Voss}, \bibinfo{person}{Gretchen Krueger}, \bibinfo{person}{Tom Henighan}, \bibinfo{person}{Rewon Child}, \bibinfo{person}{Aditya Ramesh}, \bibinfo{person}{Daniel~M. Ziegler}, \bibinfo{person}{Jeffrey Wu}, \bibinfo{person}{Clemens Winter}, \bibinfo{person}{Christopher Hesse}, \bibinfo{person}{Mark Chen}, \bibinfo{person}{Eric Sigler}, \bibinfo{person}{Mateusz Litwin}, \bibinfo{person}{Scott Gray}, \bibinfo{person}{Benjamin Chess}, \bibinfo{person}{Jack Clark}, \bibinfo{person}{Christopher Berner}, \bibinfo{person}{Sam McCandlish}, \bibinfo{person}{Alec Radford}, \bibinfo{person}{Ilya Sutskever},
  {and} \bibinfo{person}{Dario Amodei}.} \bibinfo{year}{2020}\natexlab{}.
\newblock \showarticletitle{Language models are few-shot learners}. In \bibinfo{booktitle}{\emph{Proceedings of the 34th International Conference on Neural Information Processing Systems}} (<conf-loc>, <city>Vancouver</city>, <state>BC</state>, <country>Canada</country>, </conf-loc>) \emph{(\bibinfo{series}{NIPS '20})}. \bibinfo{publisher}{Curran Associates Inc.}, \bibinfo{address}{Red Hook, NY, USA}, Article \bibinfo{articleno}{159}, \bibinfo{numpages}{25}~pages.
\newblock
\showISBNx{9781713829546}


\bibitem[Bubeck et~al\mbox{.}(2023)]%
        {bubeck2023sparks_of_agi}
\bibfield{author}{\bibinfo{person}{S{\'e}bastien Bubeck}, \bibinfo{person}{Varun Chandrasekaran}, \bibinfo{person}{Ronen Eldan}, \bibinfo{person}{Johannes Gehrke}, \bibinfo{person}{Eric Horvitz}, \bibinfo{person}{Ece Kamar}, \bibinfo{person}{Peter Lee}, \bibinfo{person}{Yin~Tat Lee}, \bibinfo{person}{Yuanzhi Li}, \bibinfo{person}{Scott Lundberg}, {et~al\mbox{.}}} \bibinfo{year}{2023}\natexlab{}.
\newblock \showarticletitle{Sparks of artificial general intelligence: Early experiments with gpt-4}.
\newblock \bibinfo{journal}{\emph{arXiv preprint arXiv:2303.12712}} (\bibinfo{year}{2023}).
\newblock


\bibitem[Cheung et~al\mbox{.}({[n.\,d.]})]%
        {cheunglarge_amplify_human_biases}
\bibfield{author}{\bibinfo{person}{Vanessa Cheung}, \bibinfo{person}{Maximilian Maier}, {and} \bibinfo{person}{Falk Lieder}.} \bibinfo{year}{[n.\,d.]}\natexlab{}.
\newblock \showarticletitle{Large Language Models Amplify Human Biases in Moral Decision-Making}.
\newblock  (\bibinfo{year}{[n.\,d.]}).
\newblock


\bibitem[Echterhoff et~al\mbox{.}(2024)]%
        {echterhoff2024cognitive_biases_LLMs}
\bibfield{author}{\bibinfo{person}{Jessica Echterhoff}, \bibinfo{person}{Yao Liu}, \bibinfo{person}{Abeer Alessa}, \bibinfo{person}{Julian McAuley}, {and} \bibinfo{person}{Zexue He}.} \bibinfo{year}{2024}\natexlab{}.
\newblock \showarticletitle{Cognitive bias in high-stakes decision-making with llms}.
\newblock \bibinfo{journal}{\emph{arXiv preprint arXiv:2403.00811}} (\bibinfo{year}{2024}).
\newblock


\bibitem[Gabriel(2020)]%
        {gabriel2020ai_alignment}
\bibfield{author}{\bibinfo{person}{Iason Gabriel}.} \bibinfo{year}{2020}\natexlab{}.
\newblock \showarticletitle{Artificial intelligence, values, and alignment}.
\newblock \bibinfo{journal}{\emph{Minds and machines}} \bibinfo{volume}{30}, \bibinfo{number}{3} (\bibinfo{year}{2020}), \bibinfo{pages}{411--437}.
\newblock


\bibitem[Huang and Chang(2023)]%
        {huang2023reasoning_survey}
\bibfield{author}{\bibinfo{person}{Jie Huang} {and} \bibinfo{person}{Kevin Chen-Chuan Chang}.} \bibinfo{year}{2023}\natexlab{}.
\newblock \bibinfo{title}{Towards Reasoning in Large Language Models: A Survey}.
\newblock
\newblock
\showeprint[arxiv]{2212.10403}~[cs.CL]


\bibitem[Ji et~al\mbox{.}(2023)]%
        {ji2023ai_alignment}
\bibfield{author}{\bibinfo{person}{Jiaming Ji}, \bibinfo{person}{Tianyi Qiu}, \bibinfo{person}{Boyuan Chen}, \bibinfo{person}{Borong Zhang}, \bibinfo{person}{Hantao Lou}, \bibinfo{person}{Kaile Wang}, \bibinfo{person}{Yawen Duan}, \bibinfo{person}{Zhonghao He}, \bibinfo{person}{Jiayi Zhou}, \bibinfo{person}{Zhaowei Zhang}, {et~al\mbox{.}}} \bibinfo{year}{2023}\natexlab{}.
\newblock \showarticletitle{Ai alignment: A comprehensive survey}.
\newblock \bibinfo{journal}{\emph{arXiv preprint arXiv:2310.19852}} (\bibinfo{year}{2023}).
\newblock


\bibitem[Liu et~al\mbox{.}(2024)]%
        {liu2024large}
\bibfield{author}{\bibinfo{person}{Ryan Liu}, \bibinfo{person}{Theodore~R Sumers}, \bibinfo{person}{Ishita Dasgupta}, {and} \bibinfo{person}{Thomas~L Griffiths}.} \bibinfo{year}{2024}\natexlab{}.
\newblock \showarticletitle{How do Large Language Models Navigate Conflicts between Honesty and Helpfulness?}
\newblock \bibinfo{journal}{\emph{arXiv preprint arXiv:2402.07282}} (\bibinfo{year}{2024}).
\newblock


\bibitem[Lu et~al\mbox{.}(2021)]%
        {lu2021fantastically_prompt_order}
\bibfield{author}{\bibinfo{person}{Yao Lu}, \bibinfo{person}{Max Bartolo}, \bibinfo{person}{Alastair Moore}, \bibinfo{person}{Sebastian Riedel}, {and} \bibinfo{person}{Pontus Stenetorp}.} \bibinfo{year}{2021}\natexlab{}.
\newblock \showarticletitle{Fantastically ordered prompts and where to find them: Overcoming few-shot prompt order sensitivity}.
\newblock \bibinfo{journal}{\emph{arXiv preprint arXiv:2104.08786}} (\bibinfo{year}{2021}).
\newblock


\bibitem[Macmillan-Scott and Musolesi(2023)]%
        {irrationality_survey_macmillan2023ir}
\bibfield{author}{\bibinfo{person}{Olivia Macmillan-Scott} {and} \bibinfo{person}{Mirco Musolesi}.} \bibinfo{year}{2023}\natexlab{}.
\newblock \showarticletitle{(Ir) rationality in AI: State of the Art, Research Challenges and Open Questions}.
\newblock \bibinfo{journal}{\emph{arXiv preprint arXiv:2311.17165}} (\bibinfo{year}{2023}).
\newblock


\bibitem[Macmillan-Scott and Musolesi(2024)]%
        {irrationality_macmillan2024ir}
\bibfield{author}{\bibinfo{person}{Olivia Macmillan-Scott} {and} \bibinfo{person}{Mirco Musolesi}.} \bibinfo{year}{2024}\natexlab{}.
\newblock \showarticletitle{(Ir) rationality and cognitive biases in large language models}.
\newblock \bibinfo{journal}{\emph{Royal Society Open Science}} \bibinfo{volume}{11}, \bibinfo{number}{6} (\bibinfo{year}{2024}), \bibinfo{pages}{240255}.
\newblock


\bibitem[Quine(1969)]%
        {Quine1969-QUIORA-6}
\bibfield{editor}{\bibinfo{person}{Willard Van~Orman Quine}} (Ed.). \bibinfo{year}{1969}\natexlab{}.
\newblock \bibinfo{booktitle}{\emph{Ontological Relativity and Other Essays}}.
\newblock \bibinfo{publisher}{Columbia University Press}, \bibinfo{address}{New York}.
\newblock


\bibitem[Rane et~al\mbox{.}(2024)]%
        {rane2024concept_align}
\bibfield{author}{\bibinfo{person}{Sunayana Rane}, \bibinfo{person}{Polyphony~J. Bruna}, \bibinfo{person}{Ilia Sucholutsky}, \bibinfo{person}{Christopher Kello}, {and} \bibinfo{person}{Thomas~L. Griffiths}.} \bibinfo{year}{2024}\natexlab{}.
\newblock \bibinfo{title}{Concept Alignment}.
\newblock
\newblock
\showeprint[arxiv]{2401.08672}~[cs.LG]


\bibitem[Schaeffer et~al\mbox{.}(2024)]%
        {schaeffer2024are_emergent_ab_mirage}
\bibfield{author}{\bibinfo{person}{Rylan Schaeffer}, \bibinfo{person}{Brando Miranda}, {and} \bibinfo{person}{Sanmi Koyejo}.} \bibinfo{year}{2024}\natexlab{}.
\newblock \showarticletitle{Are emergent abilities of large language models a mirage?}
\newblock \bibinfo{journal}{\emph{Advances in Neural Information Processing Systems}}  \bibinfo{volume}{36} (\bibinfo{year}{2024}).
\newblock


\bibitem[Touvron et~al\mbox{.}(2023)]%
        {llama2}
\bibfield{author}{\bibinfo{person}{Hugo Touvron}, \bibinfo{person}{Louis Martin}, \bibinfo{person}{Kevin Stone}, \bibinfo{person}{Peter Albert}, \bibinfo{person}{Amjad Almahairi}, \bibinfo{person}{Yasmine Babaei}, \bibinfo{person}{Nikolay Bashlykov}, \bibinfo{person}{Soumya Batra}, \bibinfo{person}{Prajjwal Bhargava}, \bibinfo{person}{Shruti Bhosale}, {et~al\mbox{.}}} \bibinfo{year}{2023}\natexlab{}.
\newblock \showarticletitle{Llama 2: Open foundation and fine-tuned chat models}.
\newblock \bibinfo{journal}{\emph{arXiv preprint arXiv:2307.09288}} (\bibinfo{year}{2023}).
\newblock


\bibitem[Tversky(1977)]%
        {Tversky1977FeaturesOS}
\bibfield{author}{\bibinfo{person}{Amos Tversky}.} \bibinfo{year}{1977}\natexlab{}.
\newblock \showarticletitle{Features of Similarity}.
\newblock \bibinfo{journal}{\emph{Psychological Review}}  \bibinfo{volume}{84} (\bibinfo{year}{1977}), \bibinfo{pages}{327--352}.
\newblock
\urldef\tempurl%
\url{https://api.semanticscholar.org/CorpusID:9173202}
\showURL{%
\tempurl}


\bibitem[Tversky and Gati(1978)]%
        {Tversky_gati_similarity}
\bibfield{author}{\bibinfo{person}{Amos Tversky} {and} \bibinfo{person}{Itamar Gati}.} \bibinfo{year}{1978}\natexlab{}.
\newblock \showarticletitle{Studies of Similarity}.
\newblock In \bibinfo{booktitle}{\emph{Cognition and Categorization}}, \bibfield{editor}{\bibinfo{person}{Eleanor Rosch} {and} \bibinfo{person}{Barbara Lloyd}} (Eds.). \bibinfo{publisher}{Lawrence Elbaum Associates}, \bibinfo{pages}{1--1978}.
\newblock


\bibitem[Wei et~al\mbox{.}(2022a)]%
        {wei2022emergent_abilities_llms}
\bibfield{author}{\bibinfo{person}{Jason Wei}, \bibinfo{person}{Yi Tay}, \bibinfo{person}{Rishi Bommasani}, \bibinfo{person}{Colin Raffel}, \bibinfo{person}{Barret Zoph}, \bibinfo{person}{Sebastian Borgeaud}, \bibinfo{person}{Dani Yogatama}, \bibinfo{person}{Maarten Bosma}, \bibinfo{person}{Denny Zhou}, \bibinfo{person}{Donald Metzler}, {et~al\mbox{.}}} \bibinfo{year}{2022}\natexlab{a}.
\newblock \showarticletitle{Emergent abilities of large language models}.
\newblock \bibinfo{journal}{\emph{arXiv preprint arXiv:2206.07682}} (\bibinfo{year}{2022}).
\newblock


\bibitem[Wei et~al\mbox{.}(2022b)]%
        {wei2022chain_of_thought}
\bibfield{author}{\bibinfo{person}{Jason Wei}, \bibinfo{person}{Xuezhi Wang}, \bibinfo{person}{Dale Schuurmans}, \bibinfo{person}{Maarten Bosma}, \bibinfo{person}{Fei Xia}, \bibinfo{person}{Ed Chi}, \bibinfo{person}{Quoc~V Le}, \bibinfo{person}{Denny Zhou}, {et~al\mbox{.}}} \bibinfo{year}{2022}\natexlab{b}.
\newblock \showarticletitle{Chain-of-thought prompting elicits reasoning in large language models}.
\newblock \bibinfo{journal}{\emph{Advances in neural information processing systems}}  \bibinfo{volume}{35} (\bibinfo{year}{2022}), \bibinfo{pages}{24824--24837}.
\newblock


\bibitem[Zhu and Griffiths(2024)]%
        {zhu2024incoherent}
\bibfield{author}{\bibinfo{person}{Jian-Qiao Zhu} {and} \bibinfo{person}{Thomas~L Griffiths}.} \bibinfo{year}{2024}\natexlab{}.
\newblock \showarticletitle{Incoherent Probability Judgments in Large Language Models}.
\newblock \bibinfo{journal}{\emph{arXiv preprint arXiv:2401.16646}} (\bibinfo{year}{2024}).
\newblock


\end{thebibliography}
\appendix
\section{Appendix}

\begin{figure*}[h]
        \vspace{-3mm}

    \includegraphics[scale = 0.6, angle=0]{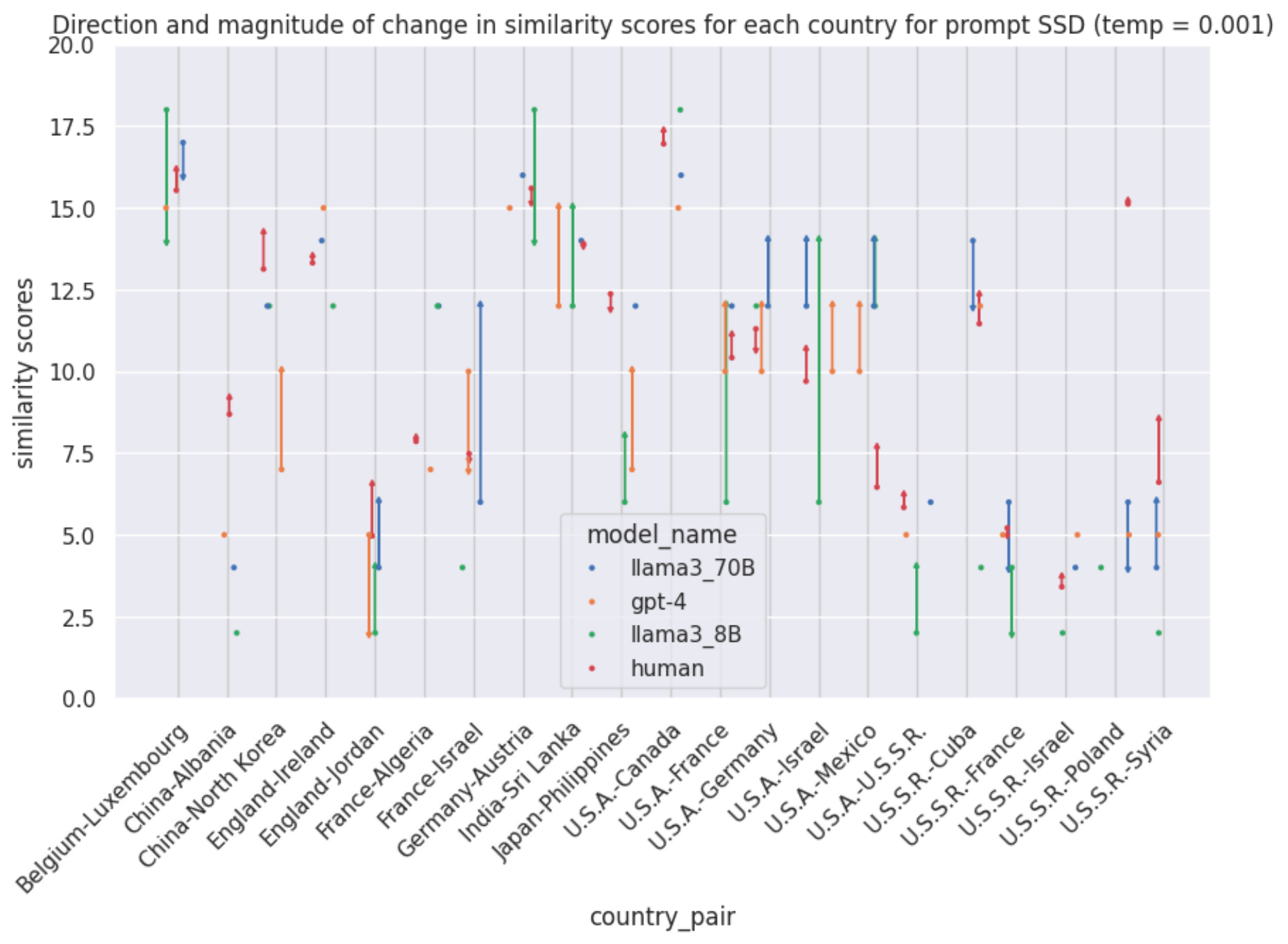}
    \caption{Comparing direction and magnitude of similarity scores for each country pair for aligned LLMs and human data (Prompt style SSD)}
    \label{fig:direction_magnitude_SSD_human}
        \vspace{-3mm}

\end{figure*}

\begin{figure*}[h]
  \centering
    \subfigure[]{\includegraphics[scale=0.27]{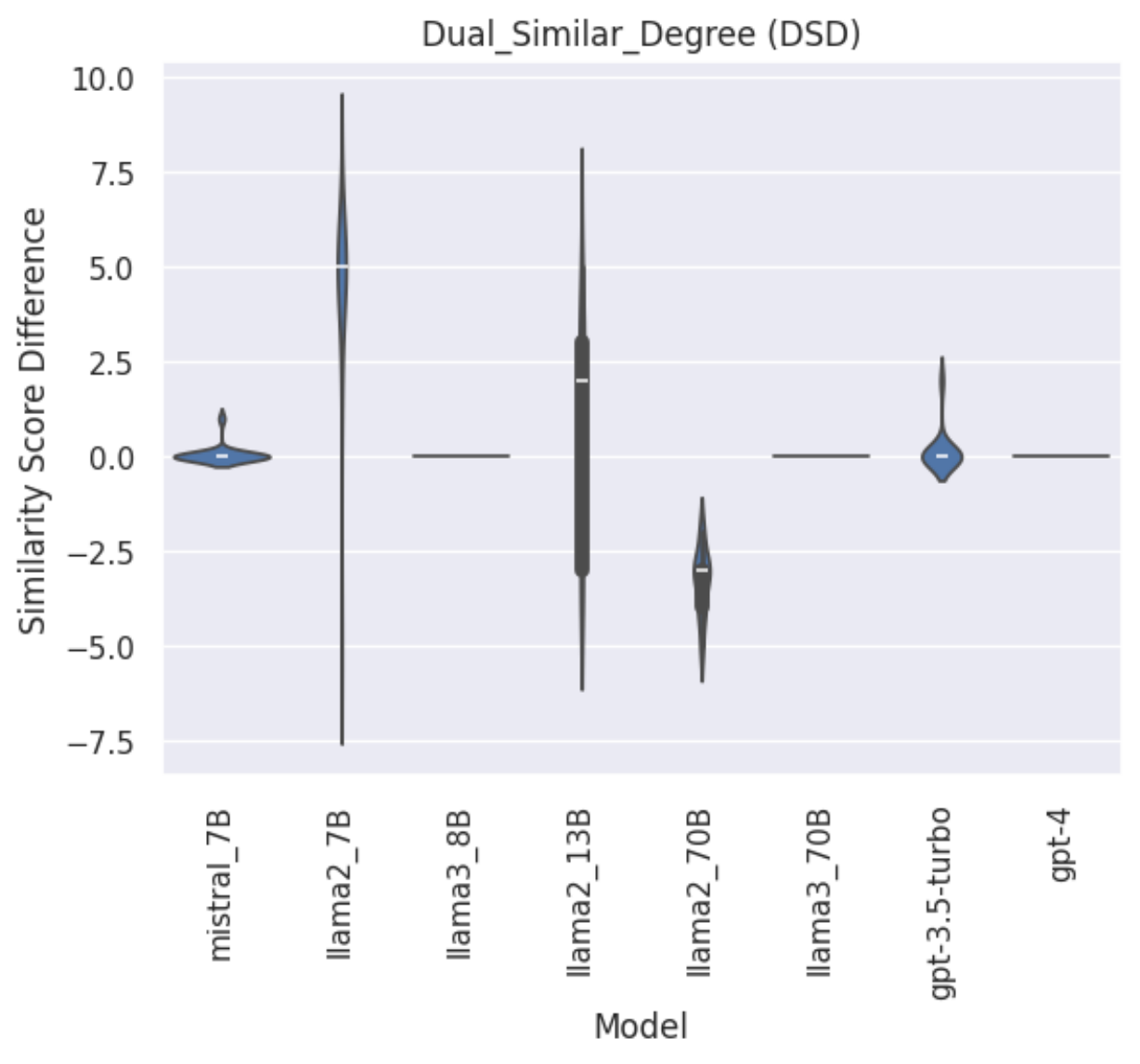}}
  \subfigure[]{\includegraphics[scale=0.27]{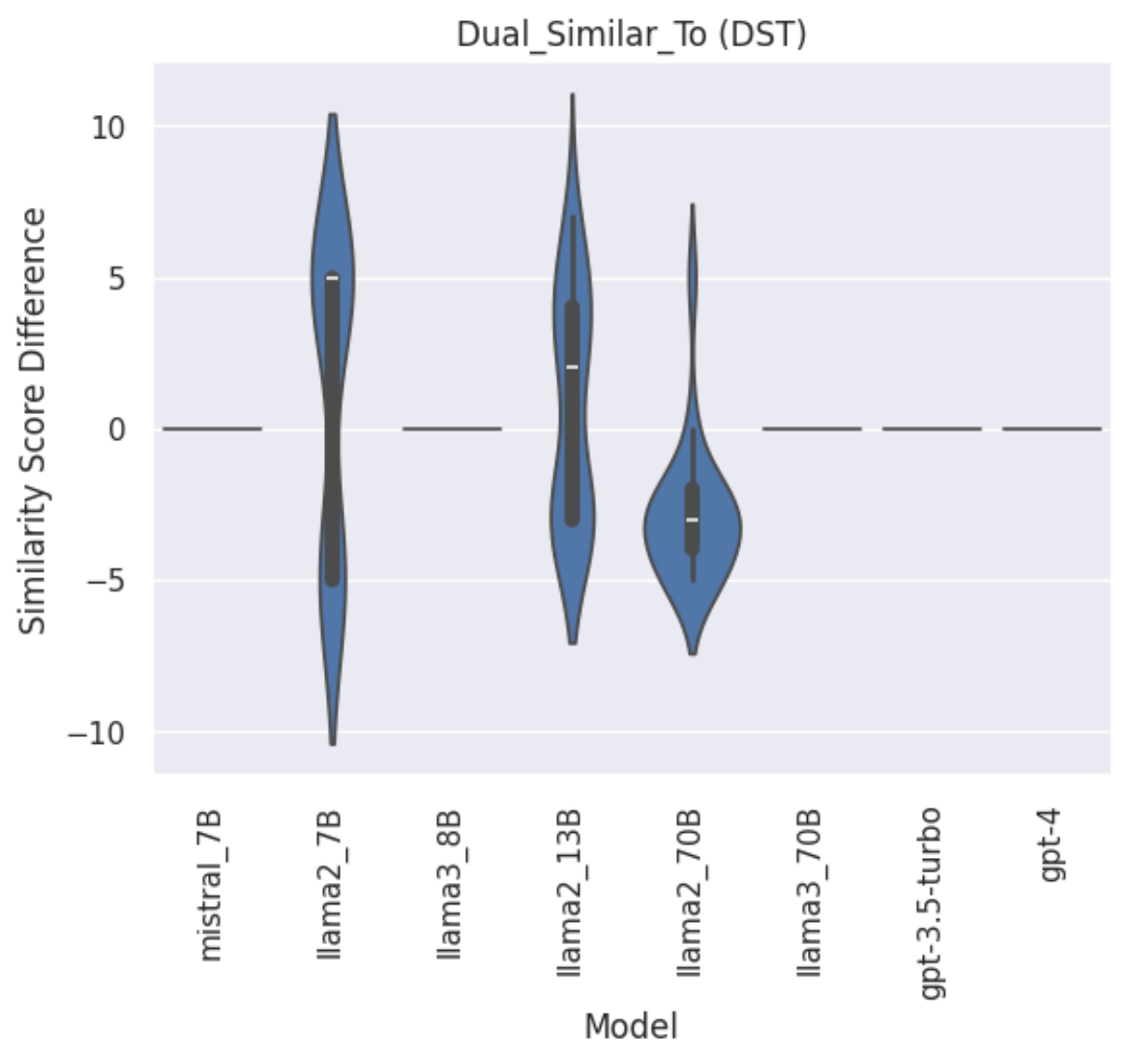}}
    \subfigure[]{\includegraphics[scale=0.27]{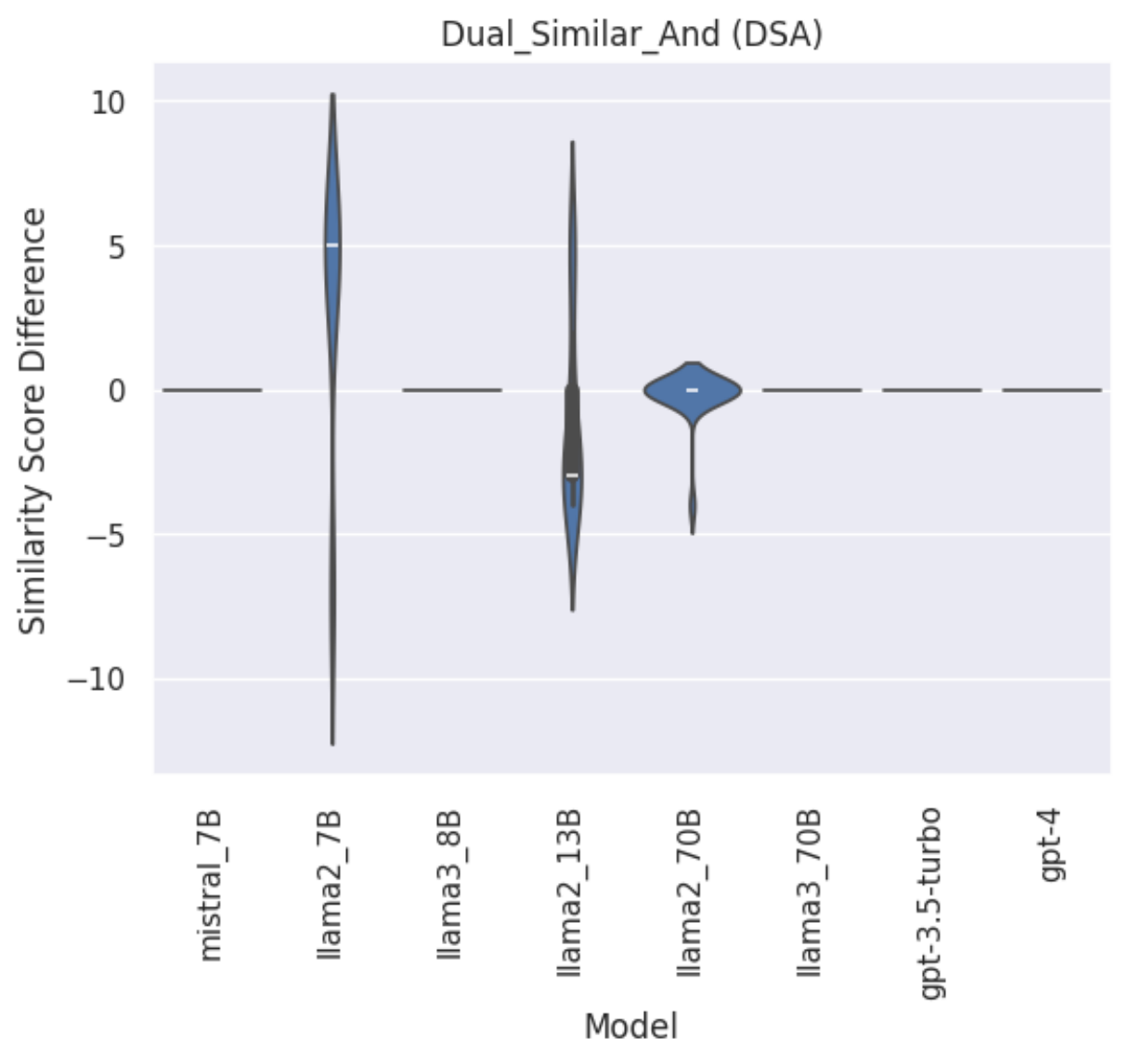}}
    \caption{Distribution of similarity differences for all countries and all models for dual prompt settings}
\label{fig:distribution-plot-dual}
        \vspace{-2mm}

\end{figure*}

\end{document}